%% file: iclr2024_conference.tex
\documentclass{article} 
\usepackage{iclr2024_conference,times}

\input{math_commands.tex}

\usepackage{hyperref}
\usepackage{url}

\usepackage{graphicx}
\usepackage{amsmath}
\usepackage{amssymb}
\usepackage{booktabs}
\usepackage{mathtools}

\usepackage{soul}
\usepackage{multicol}
\usepackage{makecell}
\usepackage[normalem]{ulem}
\usepackage{multirow}
\usepackage{booktabs}
\usepackage{color, colortbl}
\usepackage[ruled, vlined]{algorithm2e}
\definecolor{Gray}{gray}{0.9}
\usepackage{lipsum}
\usepackage{subcaption}

\usepackage{pifont}
\newcommand{\eg}{e.g., }

\newcommand{\xmark}{\ding{55}}

\newcommand{\T}{^{\intercal}}

\title{Learning Unorthogonalized Matrices for\\ Rotation Estimation}

\author{Kerui Gu\textsuperscript{1}, Zhihao Li\textsuperscript{2}, Shiyong Liu\textsuperscript{2}, Jianzhuang Liu\textsuperscript{2}, Songcen Xu\textsuperscript{2}, Youliang Yan\textsuperscript{2}, \\
\textbf{Michael Bi Mi\textsuperscript{3}, Kenji Kawaguchi\textsuperscript{1}, Angela Yao\textsuperscript{1}}\\
\textsuperscript{1}National University of Singapore\\
\textsuperscript{2}Huawei Noah’s Ark Lab\\
\textsuperscript{3}Huawei International Pte Ltd\\
\texttt{\{keruigu,ayao\}comp.nus.edu.sg}
}

\iclrfinalcopy 
\begin{document}

\maketitle

\begin{abstract}
Estimating 3D rotations is a common procedure for 3D computer vision. The accuracy depends heavily on the rotation representation. 
One form of representation -- rotation matrices -- is popular due to its continuity, especially for pose estimation tasks. The learning process usually incorporates orthogonalization to ensure orthonormal matrices. Our work reveals, through gradient analysis, that common orthogonalization procedures based on the Gram-Schmidt process and singular value decomposition will slow down training efficiency. 
To this end, we advocate removing orthogonalization from the learning process and learning unorthogonalized `\textbf{P}seudo' \textbf{Ro}tation \textbf{M}atrices (PRoM).
An optimization analysis shows that PRoM converges faster and to a better solution. By replacing the orthogonalization incorporated representation with our proposed PRoM in various rotation-related tasks, we achieve state-of-the-art results on large-scale benchmarks for human pose estimation.
\end{abstract}

\section{Introduction}

Estimating 3D rotation is a common procedure in geometry-related tasks such as 3D pose estimation.
A 3D rotation is defined by three parameters and some rotation representations like axis angles or Euler angles are specified by three parameters.  Other representations are over-parameterized, like quaternions (four) and rotation matrices (nine) and therefore must fulfill certain constraints; for example, matrices $\rmR$ in rotation groups $SO(n)$ must be orthogonal. 

Accurately estimating rotation parameters can be challenging; one reason is that some representations (axis-angles, Euler angles, or quaternions) are discontinuous~\citep{grassia1998practical,saxena09}. Recently, \citep{zhou2019continuity} proposed a continuous 6-parameter representation for 3D rotations.  The so-called `6D representation' simply drops the last column vector of the rotation matrix; the full matrix can be recovered via a Gram-Schmidt-like process.
Rotations estimated as 6D representations are more accurate than
Euler angles and quaternions~\citep{zhou2019continuity}. As such, the 6D representation has become widely adopted for human pose and shape estimation tasks.

Upon closer examination, we observe that common orthogonalization procedures, including the Gram-Schmidt process and SVD, are problematic for neural network training. Specifically, orthogonalization makes updates with standard first-order gradient descent ambiguous, thereby adding to the learning difficulty and reducing training efficiency. An explicit derivation shows that the gradient can be decomposed into potentially conflicting terms, i.e. in opposing directions, especially at the early stages of training. This finding is related not to numerical instabilities but to the internal operations for orthogonalizations like orthogonal projection and cross-product. In addition, incorporating orthogonalizations may introduce extremely large gradients that destabilize training.

To relieve these issues, we advocate removing orthogonalizations from the learning process.  Instead, we propose to learn {\textbf{P}seudo} \textbf{Ro}tation \textbf{M}atrices (PRoM), or unorthogonalized rotation matrices. Orthogonalization is applied post-hoc during inference.
A key advantage of PRoM is that the update of each matrix element is based only on the difference with respect to its corresponding ground truth; this fully avoids ambiguity and numerical instability. It also ensures that the prediction of each matrix element remains independent, as opposed to being coupled in the orthogonalization process.
PRoM converges faster than methods that keep the orthogonalization and is guaranteed to converge to a better solution, due to the non-local-injectivity of orthogonalization functions. Simply put, there are multiple estimates corresponding to the same orthogonalized matrix; this is problematic for training, which PRoM is able to avoid.

Estimated 3D rotations are often only intermediate outputs that are then applied downstream. In such cases, the learning tends to be end-to-end, with supervision from the downstream task. For example, in 3D human pose estimation, rotations are used for forward kinematics, but the supervision comes as 3D body poses.
To ensure valid rotations, existing works~\citep{zhou2019continuity, levinson2020analysis} 
orthogonalize estimated rotation matrices during both training and inference. In this scenario, we break the convention and recommend passing unorthogonalized matrices to the downstream tasks for end-to-end learning; orthogonalization is applied only as post-processing during inference.  The validity of our gradient analysis and optimization holds even regardless of whether orthogonalization is in the middle or at the end of the computational graph.
Empirically, we show that integrating PRoM into body/hand pose estimation and point cloud tasks leads to faster convergence for training and better performance for the downstream task.  Summarizing our contributions:

\begin{itemize}
    \item We uncover ambiguous updates and explosive gradients when incorporating orthogonalization into network training. To mitigate, we recommend removing orthogonalization from learning and representing the rotation with a \emph{`pseudo'} rotation matrix (PRoM).
    \item We show, via derivation, why PRoM converges faster and to a better solution than pipelines with orthogonalization, due to the {non-local-injectivity} of orthogonalization.
    \item We empirically demonstrate the superiority of PRoM on several real-world tasks with different combinations of supervision. By changing a few lines of code, we achieve state-of-the-art results on several large-scale benchmarks.

\end{itemize}

\section{Related Work}

\noindent
\textbf{Learning for rotations.} 3D rotations can be described using 3 Euler angles, axis-angles, or quaternions~\citep{rieger2004systematic}.However, several works~\citep{grassia1998practical, rieger2004systematic,saxena09, knutsson2011representing,zhou2019continuity} have pointed out that the parameterization of 3D rotations with three or four parameters is discontinuous and non-ideal for learning. To address discontinuities, \citep{zhou2019continuity} proposed a continuous 6D representation that drops the last column of the full $3\times3$ rotation matrix.  For recovery, they apply a Gram-Schmidt-like process with a cross-product operation.
In a similar use of matrices and orthogonalizations, \citep{levinson2020analysis} recommended using SVD-based orthogonalization as it is a better approximation than Gram-Schmidt process under Gaussian noise. 
Our paper focuses on rotation matrices from the perspective of learning and optimization. Our analysis proves that removing orthogonalization during learning benefits convergence.

\noindent
\textbf{3D human body/hand pose and shape estimation}. 3D rotations are critical intermediate representations for downstream tasks such as body/hand pose and shape estimation. The accuracy of predicted 3D rotations largely influences the quality of 3D mesh. The classic HMR~\citep{kanazawa2018end} adopted axis-angle rotation representation, while most subsequent works~\citep{kolotouros2019learning, kocabas2020vibe, choi2021beyond, li2022cliff} used the continuous 6D representation. Previous works mainly focus on using different supervision sources (2D and 3D keypoint locations) or network design. However, by only changing the rotation representation from 6D to PRoM, we achieve state-of-the-art results.

\begin{figure}
    \centering
    \includegraphics[width=0.98\textwidth]{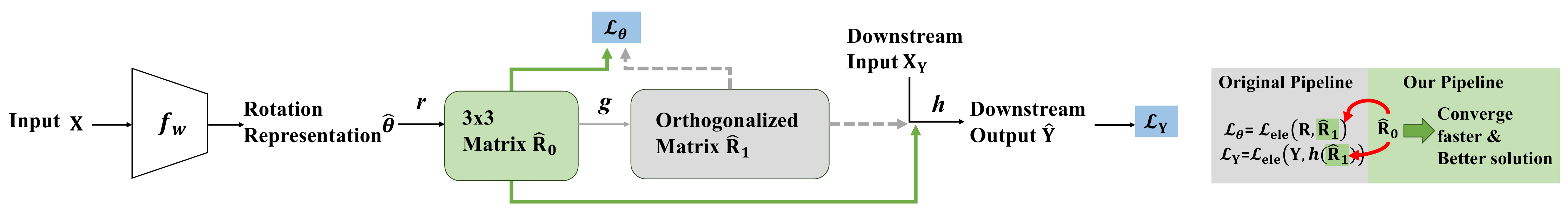}
    \caption{Rotation learning with neural networks. Existing works incorporate orthogonalization in either $r$ or $g$ to obtain the orthogonalized matrix $\hat{\rmR}_1$. We advocate directly using the unorthogonalized matrix $\hat{\rmR}_0$ for because it leads to faster convergence and better solutions.}
    \label{fig:pipeline}
\end{figure}

\section{An Analysis on Orthogonalization} 

\subsection{Preliminaries}\label{subsec:orthogonalization}

Consider a representation $\theta$ for an $n$-dimensional rotation.  $\theta$ is estimated by a neural network $f_{\vw}$, parameterized by weights $\vw$ from input $\rmX$, i.e., $\hat{\theta}=f_{\vw}(\rmX)$.  Without any assumption on the form of $\theta$, the mapping of $\theta$ to an $n\!\times\! n$ rotation matrix $\rmR\!\in\!\mathbb{R}^{n \times n}$ can be defined by $r: \theta \rightarrow \rmR$ (see Fig.~\ref{fig:pipeline}). For example, if $\theta$ is an axis-angle representation, then $r$ is the Rodrigues' rotation formula; if $\theta$ is already an $n\times n$ matrix, then $r$ is simply an identity mapping.  

Regressions with neural networks are typically unconstrained, so if the predicted $\hat{\theta}$ is {an $n \times n$} matrix, it is unlikely to be a valid rotation matrix and a subsequent orthogonalization is necessary.  We denote the unorthogonalized estimate as a ``pseudo'' rotation matrix $\rmR_0$ and the orthogonalized version as $\rmR_1$; additionally, let $g: \rmR_0 \rightarrow \rmR_1$ denote an orthogonalization procedure.

Two common orthogonalization methods are based on Singular Value Decomposition (SVD) and the Gram-Schmidt Process.  Given a matrix $\rmP\!\in\!\mathbb{R}^{n \times n}$, \textbf{Singular Value Decomposition} decomposes $\rmP$ into three matrices, i.e. $\rmP=\rmU \Sigma \rmV^{\intercal}$, where $\rmU \in \mathbb{R}^{n\times n}$, $\rmV\in \mathbb{R}^{n\times n}$ are orthogonal matrices and $\Sigma \in \mathbb{R}^{n\times n}$ is a diagonal matrix with all positive values on its diagonals. The orthogonalized version of matrix $\rmP$ can then be defined as $\rmR_{\text{SVD}}= g_{\text{SVD}}(\rmP) = \rmU \rmV^\intercal.$  The \textbf{Gram-Schmidt} process sequentially projects each column vector to be orthogonal to the previous. Consider a matrix $\rmP\!\in\!\mathbb{R}^{n \times n}=[\vp_1, \ldots, \vp_i, \ldots, \vp_n]$ with column vectors $\vp_i \in \mathbb{R}^n$.  It can be orthogonalized to $\rmR_{\text{GS}}\!=\! g_{\text{GS}}(\rmP)\!=\![\vq_1, \ldots, \vq_i, \ldots, \vq_n]$ with column vectors $\vq_i \in \mathbb{R}^n$ as follows: 
\begin{equation}\label{eqn:gs}
\vq_i = 
\begin{cases}
N(\vp_1) & \text{if } i=1 \\
N(\vp_i - \sum_{j=1}^{i-1}(\vq_{j} \boldsymbol{\cdot} \vp_i) \vq_{j}) & \text{if } 2 \leq i \leq n
\end{cases},
\end{equation}
\noindent
where $N(\cdot)$ denotes a vector normalization, i.e. $N(\vp)\! = \!\frac{\vp}{|\vp|}$ and $|\vp|$ is the magnitude of the vector $\vp$.

\subsection{6D Representation for 3D Rotations}
Since $\rmR\in\mathbb{R}^{n\times n}$ is in the set of $SO(n)$, it has only $n$ degrees of freedom and can be expressed more compactly. ~\citep{zhou2019continuity} proposed an alternative representation with $n^2-n$ parameters by simply dropping the last column vector of the rotation matrix $\rmR$.  As our interest is primarily in 3D rotations, we follow \citep{zhou2019continuity} and refer to this representation as a `6D' representation, even though it is general for $n$-dimensional rotations.

For a given rotation in the 6D representation, $\theta_{\text{6D}} = [\vt^{'}_1,\vt^{'}_2]$, 
where $\vt^{'}_i\in \mathbb{R}^{3}$ are column vectors of $\theta_{\text{6D}}$, the full rotation matrix $\rmR_{\text{6D}}$ can be determined by
$\rmR_{\text{6D}} =r_{\text{GS}}(\theta_{\text{6D}}) = [\vr'_1, \vr'_2, \vr'_3].$
\noindent The column vectors $\vr'_i \in \mathbb{R}^3$ are obtained by a Gram-Schmidt-like process: 
\begin{equation}
\vr'_i = 
\begin{cases}
N(\vt'_1) & \text{if } i=1 \\
N(\vt'_2 - (\vr'_{1} \boldsymbol{\cdot} \vt'_2) \vr'_{1}) & \text{if } i=2 \\
\vr'_1 \times \vr'_2 & \text{if } i = 3. 
\end{cases}
\label{eqn:gramschmidt_forward_repr}
\end{equation}
\noindent
The difference between $r_{\text{GS}}$ (Eq.~\ref{eqn:gramschmidt_forward_repr}) and the standard Gram-Schmidt $g_{\text{GS}}$ (Eq.~\ref{eqn:gs}) is that the last column vector $\vr'_3$ is determined by the cross product of $\vr'_1$ and $\vr'_2$. Note that that Eq.~\ref{eqn:gramschmidt_forward_repr} directly guarantees that $\rmR_{\text{6D}}$ is orthogonal; this is a claimed advantage of the 6D representation. Currently, the 6D representation is widely used in human body and hand pose and shape estimation~\citep{kolotouros2019learning, li2022cliff}.

\subsection{Downstream Tasks and Learning}

In many tasks, the estimated $\hat{\theta}$ is an intermediate output; it is transformed to a rotation matrix $\hat{\rmR}_1$ for a downstream, task-specific output $\hat{\rmY}$.  Let $h: (\rmR_1, \rmX_{\rmY}) \rightarrow \hat{\rmY}$ denote the downstream computation where $\rmX_{\rmY}$ is the input of the downstream task. For example, in human body mesh recovery, $h$ is the forward kinematics 
via statistic body models like SMPL~\citep{loper2015smpl}, and $\hat{\rmY}$ is the estimated 3D mesh vertices, while $\rmX$ is the input image and $\rmX_{\rmY}$ is the estimated shape and camera parameters. Since the downstream task of mesh recovery is widely used throughout this paper, we refer the reader to additional details in Appendix~\ref{app:details_downstream}.

The functions $r$, $g$, and $h$ form a computational graph from $\theta$ to $\rmR_1$ to the target output $\hat{\rmY} \!=\!h( g (r (\theta)))$ (see Fig.~\ref{fig:pipeline}). During learning, there may be two losses: one on the rotation, $\mathcal{L}_{\theta}$, and one on the downstream tasks, $\mathcal{L}_{\rmY}$. For tasks where the intermediate rotation ground truth is unavailable, $\mathcal{L}_{\rmY}$ can also be applied as the sole form of supervision.  For tasks that end with predicting rotations, only $\mathcal{L}_{\theta}$ is considered. In the paper, we consider all three conditions for comprehensiveness. The standard practice~\citep{zhou2019continuity, levinson2020analysis} is to calculate element-wise losses $\mathcal{L}_{\text{ele}}$ for both rotation and downstream tasks:
\begin{equation}\label{eq:general_loss}
    \mathcal{L} = \mathcal{L}_{\theta}+\mathcal{L}_{\rmY}= \mathcal{L}_{\text{ele}}(\rmR, \hat{\rmR}_1)+\mathcal{L}_{\text{ele}}(\rmY, h(\hat{\rmR}_1))
\end{equation}
where $\rmR$ and $\rmY$ are the corresponding ground truth for the rotation and downstream output, respectively, when available. The loss $\mathcal{L}_{\text{ele}}$ can be an element-wise mean-squared error (MSE) or mean-average error (MAE). 

\subsection{Gradient Analysis}
\label{sec:gradients}

We consider the case of 3D rotations. The gradients of neural network weights $\vw$ from the loss $\mathcal{L}$ in Eq.~\ref{eq:general_loss} can be formulated as
\begin{equation}\label{eq:grad_pre}
    \frac{\partial\mathcal{L}}{\partial\vw} = \left(\frac{\partial \mathcal{L}_{\theta}}{\partial g(r(\theta))}+\frac{\partial \mathcal{L}_{\rmY}}{\partial h(g(r(\theta))} \nabla h\right)\nabla g \nabla r \nabla f,
\end{equation}
\noindent
where $\frac{\partial \mathcal{L}_{\theta}}{\partial g(r(\theta))}$ and $\frac{\partial \mathcal{L}_{\rmY}}{\partial h(r(f(x))} \nabla h$ are determined by the corresponding element-wise loss and the downstream estimate (\eg forward kinematics in human mesh recovery); $\nabla g$ and $\nabla r$ denote the gradients of the orthogonalization algorithm and rotation representation (\eg 6D from~\citep{zhou2019continuity}). Here, we focus on analyzing $\nabla g\nabla r$ since it depends on the rotation representation and is multiplied by $\nabla h$ in the downstream task.

In an orthogonalized matrix, the columns must be orthonormal. Methods such as Gram-Schmidt and SVD-based orthogonalization feature column-wise operations except for vector normalization. Therefore, we analyze the gradients on a column basis and study the gradients of the orthonormal columns $g(r(f_\vw(\rmX)))=\{\vr'_1, \vr'_2, \vr'_3\}$ with respect to the unorthogonalized columns or the direct output of the network $f_\vw(\rmX)=\{\vt'_1, \vt'_2, \vt'_3\}$, where $\vr'_i$ and $\vt'_i$ are the $i^{\text{th}}$ column in the corresponding matrix. Consider the gradient of the rotation loss $\mathcal{L}_{\theta}$ w.r.t. the first unorthogonalized column $\vt'_1$, which can be expressed as
\begin{equation}\label{eq:general_ortho_grad}
\small
    \frac{\partial\mathcal{L}_{\theta}}{\partial\vt'_1} = (\vr'_1  -\vr_1)^{\intercal}\frac{\partial\vr'_1}{\partial\vt'_1}+(\vr'_2-\vr_2)^{\intercal}\frac{\partial\vr'_2}{\partial\vt'_1}+(\vr'_3-\vr_3)^{\intercal}\frac{\partial\vr'_3}{\partial\vt'_1}.
\end{equation}

In the equation above, $\{\vr_1, \vr_2, \vr_3\}$ are the corresponding ground truth vectors. For the 6D representation, $f_w(\rmX)=\{\vt'_1, \vt'_2\}$ as it generates $\vr'_3$ by a cross product though this does not influence our analysis. We provide explicit derivations of Eq.~\ref{eq:general_ortho_grad} for 6D (Gram-Schmidt-like)-based orthogonalization in Appendix~\ref{app:6d_grad}.

\subsubsection{Update Ambiguity}
\label{sec:grad_ambiguity}

We assume that the shortest path to update $\vr'_1$ to $\vr_1$ is consistent with $(\vr'_1  -\vr_1)^{\intercal}\frac{\partial\vr'_1}{\partial\vt'_1}$. Yet SGD-based optimization makes updates according to Eq.~\ref{eq:general_ortho_grad}.  Specifically, the gradient for $t'_1$ is a weighted sum of the differences between \emph{all} the orthonormal columns and their respective ground truths, i.e. with additional terms 
$(\vr'_2  -\vr_2)^{\intercal}\frac{\partial\vr'_2}{\partial\vt'_2}$ and 
$(\vr'_3  -\vr_3)^{\intercal}\frac{\partial\vr'_3}{\partial\vt'_3}$.
These extra terms are non-negligible and may give opposite directions compared to $(\vr'_1  -\vr_1)^{\intercal}\frac{\partial\vr'_1}{\partial\vt'_1}$. In short, this is an \emph{ambiguous update} and results in lower learning efficiency, especially at the beginning of training. More details on the ambiguous update are provided in Appendix~\ref{app:update_ambiguity} and empirical verification are in Sec.~\ref{sec:empirical_verifications}.

One may contend that the training speed is inconsequential as long as it ultimately yields optimal outcomes.   Sec.~\ref{sec:optimization_analysis} proves why
including orthogonalization may also lead to suboptimal results.

\begin{figure}
    \centering
    \includegraphics[width=0.98\textwidth]{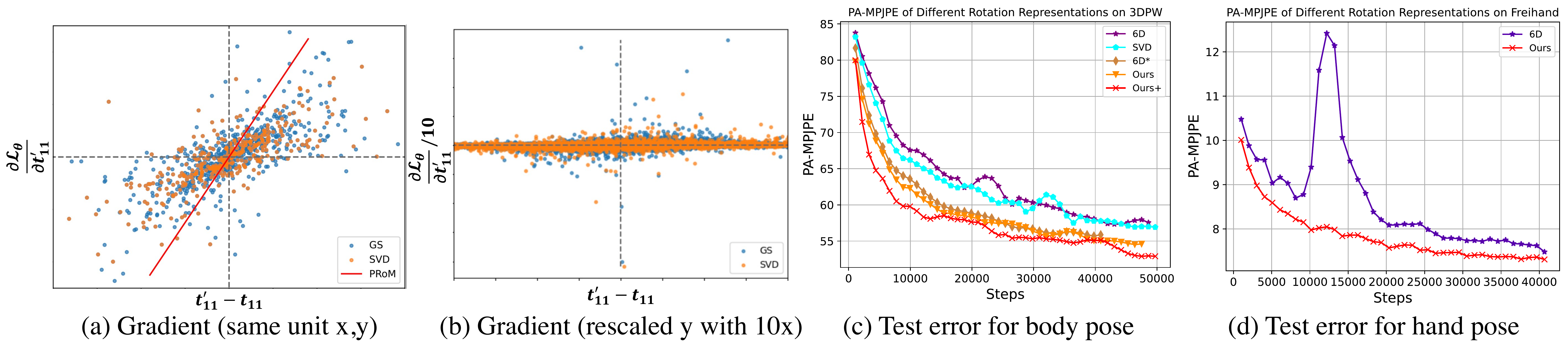}
    \caption{(a) Gradients w.r.t. $(t'_{11}-t_{11})$. PRoM shows consistent gradients for any given x, whereas 6D- and SVD-based methods show ambiguous updates with diverse values for the same x. (b) Rescaling the y-axis by a factor of 10 reveals that GS and SVD also yield very large exploding graidents.
    (c) Test error on 3DPW with different rotation representations. $\text{6D}^*$ means removing orthogonalizations during the learning of 6D representation. Ours+ indicates using a larger learning rate. (d) Test error on FreiHAND with different rotation representations. The instability of 6D gradients causes a spike in error during training.}
    \label{fig:combined_figure}
\end{figure}

\subsubsection{Gradient Explosion}
\label{sec:grad_explosion}
Orthogonalization operations like Gram-Schmidt and SVD are also prone to gradient explosion. For example, in Gram-Schmidt-based orthogonalizations, $\frac{\partial\vr'_2}{\partial\vt'_1}$ may become extremely large. From Eqs.~\ref{eqn:gs} and \ref{eqn:gramschmidt_forward_repr}, if we denote $\vr^{''}_2 = \vt'_2-(\vr'_1\cdot\vt'_2)\vr'_1$, then $\frac{\partial\vr'_2}{\partial\vt'_1}$ can be written as
\begin{equation}
\small
    \frac{\partial \vr'_2}{\partial \vt'_1}=\frac{\partial \vr'_2}{\partial \vr''_2}\frac{\partial \vr''_2}{\partial \vt'_1}=\nabla N(\vr''_2)\frac{\partial \vr''_2}{\partial \vt'_1}=-\frac{1}{|\vr''_2|}(\rmI - \frac{\vr''_2(\vr''_2)^{\intercal}}{|\vr''_2|^2})\frac{\partial \vr''_2}{\partial \vt'_1},
\end{equation}
where $\nabla N(\cdot)$ is the gradient of normalization function. When $\vt'_1$ and $\vt'_2$ are parallel, $|\vr''_2|$ approaches 0, leading to large gradients that destabilize backpropagation. Detailed derivations are shown in Appendix~\ref{app:6d_grad}. 

With SVD, the gradients depend on a matrix $\rmK$.  The matrix elements $\rmK_{ij} = \frac{1}{\lambda_i-\lambda_j}$ when $i \neq j$ and 0 otherwise, where $\lambda_i$ and $\lambda_j$ denote the $i^{\text{th}}$ and $j^{\text{th}}$ eigenvalues of the matrix being orthogonalized~\citep{ionescu2015matrix}.  This leads to either zero components or, when $\lambda_i$ is close to $\lambda_j$, very large components that result in gradient explosion~\citep{wang2021robust}.

Empirically, we observe large gradients in the synthetic and real-world experiments when incorporating the Gram-Schmidt process or SVD for orthogonalization (see details in Fig.~\ref{fig:combined_figure} (b)).  From a practical perspective, the instability can be managed with a small learning rate and a larger batch size, but this slows down the learning process.

\section{Method}

\subsection{Pseudo Rotation Matrix (PRoM)}

Given that orthogonalization may cause update ambiguities and destabilize training, we advocate representing rotations with a \emph{'pseudo'-}rotation matrix (PRoM) $\rmR_0$.  In PRoM, we simply remove the orthogonalization function $g$ (by treating it as an identity function) and apply the estimated orthogonalized matrix $\hat{\rmR}_0$ in place of the orthogonalized $\hat{\rmR}_1$, so the loss from Eq.~\ref{eq:general_loss} simplifies to 
\begin{equation}\label{eq:mseloss_unorth}
    \mathcal{L}' = \mathcal{L}'_{\theta}+\mathcal{L}'_{\rmY}=\mathcal{L}_{\text{ele}}(\rmR, \hat{\rmR}_0) + \mathcal{L}_{\text{ele}}(\rmY, h(\hat{\rmR}_0)).   
\end{equation}
With both $r(\cdot)$ and $g(\cdot)$ as identity mappings, the gradients in Eq.~\ref{eq:grad_pre} become
\begin{equation}\label{eq:grad_cur}
    \frac{\partial\mathcal{L}'}{\partial\vw} = \left(\frac{\partial \mathcal{L}'_{\theta}}{\partial \theta}+\frac{\partial \mathcal{L}'_{\rmY}}{\partial h(\theta)} \nabla h\right) \rmI \; \nabla f,
\end{equation}
\noindent
where
$\nabla g \nabla r$ simplifies into an identity matrix $\rmI$.  This simplification directly relieves the update ambiguity and graident instability. With more consistent and stable gradients, PRoM converges faster than when incorporating SVD or Gram-Schmidt; it can also tolerate higher learning rates up to 5 times.

\subsection{Optimization Analysis}
\label{sec:optimization_analysis}
In addition to faster convergence, PRoM converges to a better solution than when incorporating orthogonalization.
In the analysis of this section, we abuse notation and use $\hat{\rmR}_{0}, \hat{\rmR}_{1}$ to represent the unorthogonalized and orthogonalized matrices respectively of $m$ samples instead of one sample, i.e., $\hat{\rmR}_{1} \in \mathbb{R}^{m\times n^2}$ and $\hat{\rmR}_{0} \in \mathbb{R}^{m \times n^2}$. In addition, $\mathcal{R}^* \subset \mathbb{R}^{m \times n^2}$ is the set of all optimal matrices consisting of $m$ ground truth matrices. We consider an arbitrary total loss criterion $L$ that takes the output matrix $\hat{\rmR}_{1}$ for any orthogonalization-incorporated methods and $\hat{\rmR}_{0}$ for the proposed PRoM. 

\textbf{Theorem 1.} \emph{For any optimal matrix $\rmR^* \in \mathcal{R}^*$ and $i\in\{0,1\}$},
\begin{equation}
    L(\hat{\rmR}_{i}) \leq L(\rmR^*) + \epsilon C  \psi(\rmB_i), \,\,\, \text{where}\,\,\, C =E \psi(\rmD).
\end{equation}

Theorem 1 states that the loss from any estimated rotation matrix, $L(\hat{\rmR}_{i})$ converges at a rate that depends on the term $\psi(\rmB_i)$ (see Theorem 2 below).  More specifically, for Theorem 1, define $\epsilon\!=\!\text{sup}_{i\in\{0,1\}}\|(\frac{\partial L(\hat{\rmR}_{i})}{\partial \vw})^{\intercal}\|$, where $\|\cdot\|$ represents the Euclidean norm. As the number of iterations increases, $\epsilon \rightarrow 0$ for many gradient-based optimizers under mild conditions~\citep{bertsekas1997nonlinear} including stochastic gradient descent~\citep{fehrman2020convergence,lei2019stochastic,mertikopoulos2020almost}. $E$ is a constant defined as 
$E\!=\!\text{inf}_{\rmR^*\in\mathcal{R}^*}\text{sup}_{\hat{\rmR}\in \mathcal{R}}\| \rmR -\hat{\rmR}\|$, 
where $\mathcal{R}$ is the set of all $\hat{\rmR}_{i}$ encountered during training.
The term  $\psi(\rmM)$, for any real matrix $\rmM$, define $\psi(\rmM) = 1/\sqrt{\lambda_{\min}(\rmM \rmM\T)}$ if $\lambda_{\min}(\rmM \rmM\T)\neq 0$ and $\psi(\rmM)=\infty$ if $\lambda_{\min}(\rmM\rmM\T)= 0$.

In Theorem 1, matrices $\rmD$ and $\rmB_i$ are gradient matrices:
\begin{equation}
\small
\rmD=\frac{\partial \hat{\rmR}_{0}}{\partial \vw} \in \mathbb{R}^{mn^2\times d} \qquad \text{ and} \qquad \rmB_i=\frac{\partial \hat{\rmR}_{i}}{\partial \hat{\rmR}_{0}} \in \mathbb{R}^{mn^2\times mn^2}, \quad i \in \{0,1\},
\end{equation}
where $d$ is the number of parameters as $\vw\in \mathbb{R}^{d}$. Theorem 1 holds under the assumption that the training loss, e.g.,  $\vw\rightarrow L(\hat{\rmR}_{i})$, is \emph{non-convex} while the loss criterion $\hat{\rmR}_{i}\rightarrow L(\hat{\rmR}_{i})$, is differentiable and convex. These assumptions are satisfied by using neural networks with common loss criteria such as the MSE loss. This bound in Theorem 1 is tight. That is, there exist cases where the bound in Theorem 1 holds with equality. This is because its proof only takes two bounds that are known to be tight: i.e., the eigenvalue replacement and an inequality via the convexity of the loss criteria. Please see the proof for more details in Appendix~\ref{app:proof_theorem_1}.

We can ensure that $\lambda_{\text{min}}(\rmD\rmD^\intercal)\!\neq\!0$ by increasing the width of the neural network~\citep{kawaguchi2019depth, kawaguchi2022understanding}. As such, Theorem 1 shows that $L(\hat{\rmR}_{i})$ can converge to the optimal loss $L({\rmR}^*)$ at a rate specified by $\psi(\rmB_i)$. The property of $\psi(\rmB_i)$ depends on $i$, i.e., if the matrix is orthogonalized or not, and is given in Theorem 2. 
 
\textbf{Theorem 2.} For any $\hat{\rmR}_{0}$, $\psi(\rmB_0)=1$ and $\psi(\rmB_1)=\infty$ for any  $g$ that is not locally injective at $\hat{\rmR}_{0}$. 

Proof for Theorem 2 is given in Appendix~\ref{app:proof_theorem_2}. This theorem states that for any orthogonalization incorporated method, $\psi(\rmB_1)$ is infinity, which makes the term $\epsilon C\psi(\rmB_1)$ non-negligible. However, by removing the orthogonalization during the training, $\epsilon C\psi(\rmB_0)$ approaches 0.

Theorems 1 and 2 together establish that gradient descent with PRoM is faster than approaches that incorporate orthogonalization.  Furthermore, using PRoM can make the loss converge to the optimal value, whereas the orthogonalization incorporated methods with $i\!=\!1$ may fail to yield optimal outcomes since $\psi(\rmB_1)=\! \infty$. The difference between the two arises from the non-local-injectivity of $g$, the orthogonalization procedure (see \ref{app:1} for a simple illustration).  By removing the orthogonalization from the computational graph, as we have done with PRoM, the gradient descent is proven to find the optimal solution.

For downstream tasks, we arrive at a similar conclusion that $L(\hat{\rmY}_0)$ will converge to the optimal results whereas $L(\hat{\rmY}_1)$ may fail to do so. Proof is given in Appendix~\ref{app:proof_downstream}.

\section{Experiments}

\subsection{Empirical Verification}
\label{sec:empirical_verifications}
\noindent
\textbf{Update Ambiguity and Gradient Explosion.} We compare the gradients from the Gram-Schmidt and SVD orthogonalizations with the  direct MSE loss gradients on PRoM. For clarity, we only visualize the first element of $\vt'_1$, i.e., $t'_{11}$, and plot the gradient value w.r.t. the difference between $t'_{11}$ and its ground truth $t_{11}$. We randomly generate a ground truth rotation matrix $\rmR$ and use as the predicted rotation matrix $\hat{\rmR}$ with ${t'_{ij}}=r_{ij} + \mathcal{N}(0, \sigma^2)$ following~\citep{levinson2020analysis}. We run 10K iterations for 
$\sigma$ = 0.5 to show the distribution of the gradients on $t'_{11}$. The full results are in Appendix~\ref{app:full_gradient_ambiguity}. 

Fig.~\ref{fig:combined_figure} (a) and (b) plots, for  $\sigma\!=\!0.5$, the gradients $\frac{\partial\mathcal{L}_{\theta}}{\partial t'_{11}}$ with respect to the difference $(t'_{11}-t_{11})$. In Fig.~\ref{fig:combined_figure} (a), the x- and y-axis are of the same scale while Fig.~\ref{fig:combined_figure} (b)'s y-axis is scaled by a factor of 10.
Fig.~\ref{fig:combined_figure} (a) shows that the gradients from PRoM are consistent and fall neatly onto the diagonal.  In contrast,
incorporating Gram-Schmidt or SVD yields diverse gradients in all four quadrants of the plot; in fact, some gradients are even in the opposite direction as $(t'_{11}-t_{11})$ (i.e. points in the second and fourth quadrants).
Fig.~\ref{fig:combined_figure} (b), by increasing the scale of the y-axis, shows many outlier points that indicate extremely large (exploded) gradients.

\noindent
\textbf{Minimum Eigenvalues of $\rmB_i\rmB_i^\intercal$} is the critical difference between the orthogonalization incorporated methods and PRoM in the optimization analysis according to Theorem 1. It's clear that for the proposed PRoM, $\lambda_{\text{min}}(\rmB_0\rmB_0^\intercal)\!=\!1$. To empirically verify $\lambda_{\text{min}}(\rmB_1\rmB_1^\intercal)\!\rightarrow\!0$, we record the matrices during the training of pose and shape estimation tasks and calculate $\lambda_{\text{min}}(\rmB_1\rmB_1^\intercal)$. For both 6D-based and SVD-based learning, $\lambda_{\text{min}}(\rmB_1\rmB_1^\intercal)$ is smaller than 1e-18, which verifies that $\lambda_{\text{min}}(\rmB_1\rmB_1^\intercal)\!\rightarrow\!0$.

\begin{table}[!t]
\centering
 \def\arraystretch{1.1}
        \resizebox{0.98\textwidth}{!}{%
                \begin{tabular}{l|cc|ccc|cc|cc}
                    
                        \specialrule{.1em}{.05em}{.05em}
            
                        & \multicolumn{2}{c}{Human3.6M} & \multicolumn{3}{c}{3DPW}& \multicolumn{2}{c}{AGORA} & \multicolumn{2}{c}{FreiHAND} \\
                        \cmidrule(lr){2-3} \cmidrule(lr){4-6} \cmidrule(lr){7-8} \cmidrule(lr){9-10}
                        Method &MPJPE $\downarrow$ & PA-MPJPE $\downarrow$ & MPJPE $\downarrow$ & PA-MPJPE $\downarrow$  & MPVPE $\downarrow$& MPJPE $\downarrow$& MVE $\downarrow$ & PA-MPVPE $\downarrow$ & PA-MPJPE $\downarrow$ \\
                        \midrule
             MANO CNN& -& -&-&-&-& -&-&10.9&11.0\\
             SPIN & - & 41.1 & 96.9 &  59.2  & 116.4 &153.4&148.9 & - & - \\
                          VIBE&  65.9 &  41.5&  93.5 &  56.5  &  113.4& -&- & -  & -     \\
             HybrIK& 54.4& 34.5& 80.0 & 48.8  & 94.5 &77.0 &73.9 & -  &- \\
         CLIFF & 47.1& 32.7& 69.0& 43.0  & 81.2 &81.0&76.0&6.6 & 6.8  \\
                        \cmidrule(lr){1-10}
         CLIFF + PRoM & \textbf{43.8} (-3.3)&\textbf{30.4} (-2.3) & \textbf{67.6} (-1.4)& \textbf{42.0} (-1.0)  & \textbf{79.2} (-2.0) &\textbf{65.0}&\textbf{61.0} & 6.4 (-0.2) & 6.5 (-0.3)  \\
            \specialrule{.1em}{.05em}{.05em}
                \end{tabular}%
        }
        \caption{\small
        Evaluation of state-of-the-art methods on Human3.6M~\citep{ionescu2013human3}, 3DPW~\citep{von2018recovering}, AGORA~\citep{agora} and FreiHAND~\citep{zimmermann2019freihand}. We achieve the best results among all the methods on 3D human body and hand pose benchmarks.}
        \label{tab:sota}
\end{table}{}

\begin{table}[t]
\begin{flushleft}
\begin{minipage}[t]{\columnwidth}
  \begin{minipage}[b]{0.49\columnwidth}
  \def\arraystretch{1.4}
\resizebox{0.98\columnwidth}{!}{
    \begin{tabular}{l|ccc|ccc}
    \toprule
    &\multicolumn{3}{c}{rotation recovery}&\multicolumn{3}{c}{point cloud pose estimation}\\
\cmidrule(lr){2-4} \cmidrule(lr){5-7}
         & Mean(\textdegree)& Max(\textdegree) & Std(\textdegree) & Mean(\textdegree)& Max(\textdegree) & Std(\textdegree)\\
         \midrule
    Axis-Angle& 3.69 & 179.22 & 5.99&11.93&179.7&21.35\\
    Euler & 6.98 & 179.95& 17.31&14.13&179.67&23.8\\
    Quat & 3.32 & 179.93& 5.97&9.03&179.66&16.33\\
    6D & 0.49& 1.98& 0.27&2.85&179.83&9.16\\
    PRoM & \textbf{0.37} & \textbf{1.86} &  \textbf{0.22} & \textbf{2.13} & \textbf{179.53} &\textbf{7.87}\\
     \bottomrule
    \end{tabular}}
  \end{minipage}
  \hfill
  \begin{minipage}[b]{0.49\columnwidth}
  \def\arraystretch{1.4}
    \resizebox{0.95\columnwidth}{!}{
    \begin{tabular}{l|ccc}
    \toprule
         &  MPJPE $\downarrow$ & PA-MPJPE $\downarrow$ & MPVPE $\downarrow$\\
     \midrule
     SMPLify & -  & 139.5 & - \\
     SPIN & -  & 52.0&-\\
     EFT & - & 49.3 & - \\
     CLIFF*& 52.8 & 32.8 & 61.5\\  
     CLIFF* + PRoM & \textbf{49.5} (-3.2) & \textbf{29.9} (-2.9) & \textbf{56.9} (-4.6)\\
     \bottomrule
    \end{tabular}}
    \end{minipage}
\end{minipage}
\end{flushleft}
\begin{flushleft}
\begin{minipage}[b]{0.49\columnwidth}
\centering({\small a}) 
\end{minipage}
\hfill
\begin{minipage}[b]{0.49\columnwidth}
\centering({\small b})
\end{minipage}
\end{flushleft}
\caption{\small (a) Comparison of methods through the mean, maximum, standard deviation of errors of rotation recovery (left) and point cloud pose estimation test (right). Compared with traditional methods and 6D representation, our method has the smallest errors.  (b) Evaluation of optimization-based methods on 3DPW providing the 2D ground truth. CLIFF* denotes the CLIFF annotator that refines 3D rotation by 2D ground truth. Our method achieves the best result with a 7\% reduction in PA-MPJPE.}
\label{tab:comb_1}
\vspace{-0.5em}
\end{table}

\begin{table}[t]
\begin{flushleft}
\begin{minipage}[t]{\columnwidth}
  \begin{minipage}[b]{0.49\columnwidth}
  \def\arraystretch{1.4}
\resizebox{0.98\columnwidth}{!}{
    \begin{tabular}{l|cccc}
    \toprule & $\mathcal{L}_{\theta}$ & $\mathcal{L}_{\rmY}$ & post & PA-MPJPE$\downarrow$\\
     \midrule
     \cite{zhou2019continuity}   & $r_{\text{GS}}$  &$r_{\text{GS}}$ & \xmark  & 57.5 \\
       \cite{levinson2020analysis} & $g_{\text{SVD}}$ & $g_{\text{SVD}}$& \xmark & 56.7\\
       \midrule
      &id. &$r_{\text{GS}}$ & $r_{\text{GS}}$ & 57.1 \\
       &$r_{\text{GS}}$ & id. & $r_{\text{GS}}$ & 55.8\\
       & id. & id. & $r_{\text{GS}}$ & 55.6\\
       Ours & id. & id. &  $g_{\text{SVD}}$ & \textbf{54.8} \\
     \bottomrule
    \end{tabular}}
  \end{minipage}
  \hfill
  \begin{minipage}[b]{0.49\columnwidth}
  \def\arraystretch{1.4}
    \resizebox{0.95\columnwidth}{!}{\begin{tabular}{cccc}\hline
      LR & \citep{zhou2019continuity} & \citep{levinson2020analysis}& PRoM\\ \hline
        1e-4 & 58.3 & 56.7 & 54.8 \\
        2e-4 & - & - & 53.4 \\
        5e-4 & - & -&52.6\\
        8e-4 & -&-&- \\\hline
      \end{tabular}}
    \end{minipage}
\end{minipage}
\end{flushleft}
\begin{flushleft}
\begin{minipage}[b]{0.49\columnwidth}
\centering({\small a}) 
\end{minipage}
\hfill
\begin{minipage}[b]{0.49\columnwidth}
\centering({\small b})
\end{minipage}
\end{flushleft}
\vspace{-0.5em}
\caption{\small (a) Ablation study of removing orthogonalizations in only $\mathcal{L}_{\theta}$ or $\mathcal{L}_{\rmY}$. `id.' means identity mapping. We see that using identity mapping in $\mathcal{L}_{\rmY}$ is more critical. (b) Different methods with varying learning rates. `-' denotes NaN. We show that PRoM can tolerate up to 5 times larger learning rates and gives better performance.}
\label{tab:comb_2}
\vspace{-0.5em}
\end{table}

\subsection{State-of-the-Art Comparisons}

We verify the effectiveness of the proposed PRoM on various tasks based on the type of supervision:
tasks in which the objective is to predict rotations  (Sec.~\ref{sec:exp_theta_only}), tasks supervised by ground truth rotations and downstream outputs (Sec.~\ref{sec:exp_both}), and tasks supervised only by ground truth downstream labels without intermediate rotation labels (Sec.~\ref{sec:exp_downstream_only}). Note that PRoM only drops orthogonalization during training; the estimated rotation matrices are still orthogonalized during inference.
We use SVD unless otherwise noted.
Appendix~\ref{app:details_experiments} gives detailed settings for all the experiments.

\subsubsection{Rotation Supervision With $\mathcal{L}_{\theta}$}
\label{sec:exp_theta_only}
\noindent
\textbf{Recovering Rotations with a Neural Network.} We follow~\citep{zhou2019continuity} and estimate rotation recovered through an auto-encoder.
The inputs (and estimated outputs) are 3D rotation matrices generated by uniformly sampling axes and angles.
The output is evaluated by the geodesic distance between the input matrix and itself. Results in the left panel of Table~\ref{tab:comb_1} (a) show that our method has the lowest error, indicating that removing orthogonalizations brings better results compared with continuous 6D representation and traditional 3D and 4D representations.

\subsubsection{Rotation and Downstream Supervision with $\mathcal{L}_{\theta}$ and $\mathcal{L}_{\rmY}$}
\label{sec:exp_both}

3D human body and hand pose and shape estimation tasks are challenging and require estimating rotation and downstream outputs. Previous works mainly focus on network designs and additional information help. We show that by simply changing the rotation representation to the proposed PRoM, we obtain significant improvement and surpass state-of-the-art results.

For both the body and hand, we follow the network design and training settings of the state-of-the-art method CLIFF~\citep{li2022cliff}. 
Appendix~\ref{app:details_body_pse} outlines the implementation details.  
For evaluation, we consider 3D Euclidean distances in millimeters (mm) between predictions and the ground truth: \textbf{MPJPE} (Mean Per Joint Position Error) and \textbf{PA-MPJPE} (Procrustes-Aligned MPJPE) for the 3D joints,
and \textbf{PVE} (Per Vertex Error)
for the mesh vertices.
In all three metrics, lower values indicate better performance. 

\paragraph{3D Body Pose and Shape Estimation}
Following previous work \citep{kolotouros2019learning,li2022cliff,lin2021end}, we train the network with a mix of datasets, including Human3.6M~\citep{ionescu2013human3}, MPI-INF-3DHP~\citep{mehta2017monocular}, 3DPW \citep{von2018recovering}, MSCOCO~\citep{lin2014microsoft}, and MPII~\citep{andriluka20142d}, using the pseudo ground truth provided by the CLIFF annotator~\citep{li2022cliff} for 2D datasets. Evaluation is performed on the indoor dataset Human3.6M, the outdoor dataset 3DPW, and the synthetic dataset AGORA.

Table~\ref{tab:sota} compares our results with state-of-the-art model-based methods\citep{kolotouros2019learning,kocabas2020vibe,li2021hybrik,li2022cliff}. Our baseline is built on the SOTA method  CLIFF~\citep{li2022cliff}; incorporating PRoM reduces the error by 1.0 - 2.0mm for 3DPW. On Human3.6M, we achieve an impressive 2.3-3.3mm or $7\%$ reduction in error over CLIFF. We also rank 1st on the AGORA~\citep{agora} leaderboard, demonstrating the effectiveness of our approach.

\textbf{Discussion} We do 
a per-sample comparison of PRoM versus the 6D representation shows that PRoM reduces the errors for 72\% of the samples. Since PRoM improves the learning process, it is logical that it achieves general improvements statistically.
Appendix~\ref{app:vis} shows visual examples.

\paragraph{3D Hand Pose and Shape Estimation} Similar to the body case, we also build on CLIFF.  For the hand, we perform mixed-dataset training on FreiHAND~\citep{zimmermann2019freihand}, Obman~\citep{hasson2019learning}, and Interhand2.6M~\citep{moon2020interhand2}, and evaluate on FreiHAND. Table~\ref{tab:sota} shows that we achieve the lowest error among the methods.

\noindent
\textbf{Point Cloud Pose Estimation.} We verify on the point cloud pose estimation task introduced in ~\citep{zhou2019continuity}, which uses 2,290 airplane point clouds to train from ShapeNet~\citep{chang2015shapenet}. The test set consists of 400 held-out point clouds augmented with 100 random rotations. The right panel of Table.~\ref{tab:comb_1} (a) shows that PRoM is state-of-the-art, with a lower mean and max error in the estimated rotation and a smaller standard deviation.

\subsubsection{Only Downstream Supervision $\mathcal{L}_{\rmY}$}
\label{sec:exp_downstream_only}

\textbf{Body Pose and Shape Estimation with 2D Ground Truth.} Since 3D rotation and pose annotations are hard to obtain, it is common to refine predictions with 2D ground truth keypoints, especially for in-the-wild datasets like 3DPW. In this case, the task has downstream supervision. Specifically, we compare our method with optimization-based methods where the mesh vertices can be optimized using 2D ground truth keypoints and rotation matrices are intermediate outputs. In Table~\ref{tab:comb_1} (b), our method is 3.3mm better than the baseline method on PA-MPJPE. We are the first to reduce the error of PA-MPJPE to under 30mm on the 3DPW test set with only a change of rotation representation.

\subsection{Ablation Study}
\label{sec:ablation}
We perform ablations on 3D human body pose and shape estimation and report PA-MPJPE on 3DPW. The settings are all as follows if not stated otherwise. We train a CLIFF-Res50 model on MSCOCO with CLIFF pseudo-GT \citep{li2022cliff}, as it is fast to train and provides comparable performance on 3DPW. For each ablation, we train with 250 epochs.

\noindent
\textbf{Impact of removing orthogonalizations.} The core of PRoM is to remove orthogonalization during training when estimating $\mathcal{L}_{\theta}$ and $\mathcal{L}_{\rmY}$. Previous works~\citep{zhou2019continuity, levinson2020analysis} employ $r_{\text{GS}}$ and $g_{\text{SVD}}$ as orthogonalizations in both losses. The impact of removing orthogonalization is given in Table~\ref{tab:comb_2} (b); while the removal for either loss yields improvements, it is more critical to remove the orthogonalization in the downstream task, i.e. using $\mathcal{L}'_{\rmY}$ from Eq.~\ref{eq:mseloss_unorth}.  This impact has never been considered in prior work.

\noindent
\textbf{Training speed.} 
Fig.~\ref{fig:combined_figure} (c) shows the evaluations of PA-MPJPE for every 1k steps with a fixed learning rate on the 3DPW test set. PRoM consistently trains faster than methods that incorporate orthogonalizations. Fig.~\ref{fig:combined_figure} (d) shows an interesting training curve when training the hand models where the 6D method exhibits a very large jump in error, which we speculate to be due to the unstable gradients of the 6D representation.  In contrast, our method steadily reduces error in a smooth manner.

\noindent
\textbf{Learning rate.} 
We demonstrate the performance of 6D~\citep{zhou2019continuity}, SVD~\citep{levinson2020analysis}, and our method with different learning rate settings to test the gradient stability. 
The learning rate starts at 1e-4, which is common setting, and increases to 2e-4, 5e-4 and 8e-4. From Table~\ref{tab:comb_2} (b), we can see that higher learning rates than 1e-4 result in NaN when incorporating orthogonalizations. However, when increasing the learning rate, our method only explodes at an extremely high learning rate of 8e-4 and achieves significantly better results.

\noindent
\textbf{Different Models.} We show the generalization ability of our method on different human pose models, which adopt different network designs and employ 6D representation. The results are in Appendix~\ref{app:details_different_models} and show that PRoM performs better than 6D representation in all metrics.

\section{Conclusion}

We studied the gradients when incorporating orthogonalizations in the learning of rotation matrices and uncovered an ambiguous and explosive gradient issue. We therefore advocate removing orthogonalization procedures from the learning process and instead using pseudo rotation matrices (PRoM).  Theoretically, we prove that PRoM will converge faster and to a better solution. By changing a few lines of code, we demonstrate state-of-the-art results on several benchmarks.

\bibliography{iclr2024_conference}
\bibliographystyle{iclr2024_conference}

\appendix
\section{Appendix}

\subsection{Details of Downstream tasks}
\label{app:details_downstream}

Here, we provide the details of downstream tasks with the example of mesh recovery tasks, which are also known as pose and shape estimation tasks. Following~\citep{kanazawa2018end, kolotouros2019learning}, the 3D mesh of the body is generated via the Skinned Multi-Person Linear (SMPL) model~\citep{loper2015smpl}, which represents the 3D mesh by shape parameters $\beta \in \mathbb{R}^{10}$ and pose parameters $\theta \in \mathbb{R}^{3K}$ where $K$ is the number of joints. The shape parameters are coefficients of a PCA shape space. Here we focus on the pose parameters, which consist of $K$ rotations. Taking body mesh recovery as an example, the network outputs the relative 3D rotation of $K\!=\!23$ joints. HMR~\citep{kanazawa2018end} utilized axis-angle representations and therefore outputs $3\times23$ pose parameters in total. However, the subsequent works~\citep{kolotouros2019learning, li2021hybrik, li2022cliff, moon2020i2l} all applied 6D representations due to the continuity.

SMPL is an end-to-end and differentiable function that generates a triangulated mesh $\rmY_{M}$ with 6980 vertices by transforming the predefined rest template $\tilde{\mathcal{T}}$ conditioned on pose and shape parameters. We denote this transformation process as $\mathcal{M}$, i.e., $\rmY_M = \mathcal{M}(\tilde{\mathcal{T}}, \theta, \beta)$. The 3D keypoints $\rmY_{J}$ are obtained by applying a linear regression on the mesh vertices. We denote this linear transformation as $\mathcal{J}$, i.e., $\rmY_J=\mathcal{J}(\rmY_M)$. The core operation during $\mathcal{M}$ is deforming the mesh by the given rotations $\rmR$ with $\rmR=f(\theta)$, which is essentially matrix multiplication. In this case, the downstream outputs can be concluded as
\begin{equation}
    \hat{\rmY}_M = \mathcal{M}(\tilde{\mathcal{T}}, f(\hat{\theta}),\hat{\beta}), \,\,\, \rmY_J=\mathcal{J}(\mathcal{M}(\tilde{\mathcal{T}}, f(\hat{\theta}),\hat{\beta})).
\end{equation}

To ensure the feasibility of the mesh, the previous method applied Rodrigues' rotation formula or a Gram-Schmidt-like process as $f$ on axis-angle and 6D representations, respectively. However, $f(\hat{\theta})$ itself does \emph{not} necessarily need to be orthogonal matrices, but only needs to be $K\times3\times3$ matrices. Through our study, we show that during the learning process, the incorporation of orthogonalizations has a negative influence on the convergence rate and generalization ability on the test set. Also, empirically, we show that removing the orthogonalizations during the training process greatly outperforms keeping them on several large-scale datasets. Note that during inference, we still use orthogonalizations to ensure the feasibility of the mesh since backward propagation is no longer utilized.

For the hand pose and shape estimation task, the only difference is that it utilizes a different parametric model, MANO~\cite{romero2022embodied} with different numbers of pose and shape parameters, but it shares the same pipeline as body pose and shape estimation.

\subsection{Full gradients of 6D-based orthogonalization}
\label{app:6d_grad}
Recall that $r$ for the 6D representation is given by the Gram-Schmidt-like process in Eq.~\ref{eqn:gramschmidt_forward_repr}, and $g$ is an identity mapping. 
Let $\{\vt'_1, \vt'_2\}$ be the column vectors in $\theta_{\text{6D}}$, $\{\vr'_1, \vr'_2, \vr'_3\}$ be the resulting vectors after applying $f_\text{GS}$, and $\{\vr_1, \vr_2, \vr_3\}$ be the corresponding ground truth. We denote $\vr''_2$ as $(\vt'_2-(\vr'_1\boldsymbol{\cdot}\vt'_2)\vr'_1)$, which is the unnormalized value of $\vr'_2$. The gradient of $\vt'_1$ from $\mathcal{L}_{\text{6D}}$ can be given as 
\begin{align}
\!\!\!\! \frac{\partial\mathcal{L}_{\text{6D}}}{\partial \vt'_1}\! =\! (\vr'_1 & -\vr_1)^{\intercal}\frac{\partial \vr'_1}{\partial \vt'_1}+(\vr'_2-\vr_2)^{\intercal}\frac{\partial \vr'_2}{\partial \vt'_1}+(\vr'_3-\vr_3)^{\intercal}\frac{\partial \vr'_3}{\partial \vt'_1} ,\label{eq:grad_general}\\
 \text{where}\;\; \frac{\partial \vr'_1}{\partial \vt'_1} \!&=\! \nabla N(\vt'_1)\!=\!\frac{1}{|\vt'_1|}(\rmI_3 - \frac{\vt'_1 (\vt'_1)^{\intercal}}{|\vt'_1|^2}),\\
\frac{\partial \vr'_2}{\partial \vt'_1}\!&=\!\frac{\partial \vr'_2}{\partial \vr''_2}\frac{\partial \vr''_2}{\partial \vr'_1}\frac{\partial \vr'_1}{\partial \vt'_1}\!=\!\nabla N(\vr''_2)\frac{\partial\vr''_2}{\partial\vr'_1}\frac{\partial \vr'_1}{\partial \vt'_1}, \label{eq:grad_general_2}\\
&=-\frac{1}{|\vr''_2|}(\rmI_3 - \frac{\vr''_2(\vr''_2)^{\intercal}}{|\vr''_2|^2})((\vr'_1 \boldsymbol{\cdot} \vt'_2)\rmI+\vr'_1(\vt'_2)^{\intercal})\frac{\partial \vr'_1}{\partial \vt'_1}, \label{eq:grad_terms}\\
\frac{\partial \vr'_3}{\partial \vt'_1}\!&=\!\frac{\partial \vr'_3}{\partial \vr'_2}\frac{\partial \vr'_2}{\partial \vt'_1}
+\frac{\partial \vr'_3}{\partial \vr'_1}\frac{\partial \vr'_1}{\partial \vt'_1}\!=\![\vr'_1]_{\times}\frac{\partial \vr'_2}{\partial \vt'_1}-[\vr'_2]_{\times}\frac{\partial \vr'_1}{\partial \vt'_1},\label{eq:grad_general_3}
\end{align}
$\nabla N(\cdot)$ is the gradient from the vector normalization, $\rmI$ is the identity matrix, and $[\vr]_{\times}$ is the skew-symmetric matrix of vector $\vr$.

\subsection{Explanations of Update Ambiguity}
\label{app:update_ambiguity}
This claim is established based on two conditions that $(\vr'_i  -\vr_i)^{\intercal}\frac{\partial\vr'_i}{\partial\vt'_1}$ should be isotropic and non-negligible where $i\!=\!\{2,3\}$. For the isotropy, it is guaranteed by the fact that at the initial stage of training, $\vr'_i$ tends to be randomly generated and therefore $\vr_i$ can be viewed as isotropic around $\vr'_i$ together with the fixed $\frac{\partial\vr'_i}{\partial\vt'_1}$ given the neural network $f_{\vw}$ and input $\rmX$. For the non-negligibility, as the orthogonalizations only incorporate multiplications and additions, this obviously holds. We also provide explicit derivations for $(\vr'_i  -\vr_i)^{\intercal}\frac{\partial\vr'_i}{\partial\vt'_1}$ w.r.t. common orthogonalizations including Gram-Schmidt-based and SVD-based in Appendix~\ref{app:6d_grad}. Under two conditions, we can imagine a case where $\vr'_1$ are far away from its ground truth $\vr_1$ but the gradient on $\vt'_1$ is zero because of the influence from the second and third column. This will greatly lower down the training speed.

\subsection{Proof of Theorem 1}
\label{app:proof_theorem_1}
\emph{Proof.} Let $i \in \{0,1\}$. By the chain rule,
\begin{equation}
    \frac{\partial L(\hat{\rmR}_{i})}{\partial \vw} = \frac{\partial L(\hat{\rmR}_{i})}{\partial \hat{\rmR}_{i}}\frac{\partial \hat{\rmR}_{i}}{\partial \hat{\rmR}_{0}}\frac{\partial \hat{\rmR}_{0}}{\partial \vw}.
\end{equation}

\noindent
By the definition of $\epsilon$,
\begin{align}
    \epsilon^2\geq \bigg\|\biggl( \frac{\partial  L(\hat{\rmR}_{i})}{\partial\vw}\biggl) ^{\intercal}\bigg\|^2 &=\bigg\|\biggl( \frac{\partial L(\hat{\rmR}_{i})}{\partial \hat{\rmR}_{i}}\frac{\partial \hat{\rmR}_{i}}{\partial \hat{\rmR}_{0}}\frac{\partial \hat{\rmR}_{0}}{\partial \vw}\biggl)^\intercal\bigg\|^2 \\
    &= \frac{\partial L(\hat{\rmR}_{i})}{\partial \hat{\rmR}_{i}}\frac{\partial \hat{\rmR}_{i}}{\partial \hat{\rmR}_{0}}\Biggl(\frac{\partial \hat{\rmR}_{0}}{\partial \vw}\biggl(\frac{\partial \hat{\rmR}_{0}}{\partial \vw}\biggl)^\intercal\Biggl)\biggl(\frac{\partial \hat{\rmR}_{i}}{\partial \hat{\rmR}_{0}}\biggl)^\intercal\biggl(\frac{\partial L(\hat{\rmR}_{i})}{\partial \hat{\rmR}_{i}}\biggl) ^\intercal \\
    &= \frac{\partial L(\hat{\rmR}_{i})}{\partial \hat{\rmR}_{i}}\rmB_i\Biggl(\rmD\rmD^\intercal\Biggl)\rmB_i^\intercal\biggl(\frac{\partial L(\hat{\rmR}_{i})}{\partial \hat{\rmR}_{i}}\biggl) ^\intercal
\end{align}
Since $\rmD\rmD^\intercal$ is a real symmetric matrix, by the eigendecomposition of $\rmD\rmD^\intercal=\rmQ \Lambda \rmQ^\intercal$, we have 
\begin{align}
\epsilon^2 &\ge \frac{\partial L(\hat{\rmR}_{i})}{\partial \hat{\rmR}_{i}}\rmB_i \Biggl(\rmD\rmD^\intercal\Biggl)\rmB_i^\intercal\biggl(\frac{\partial L(\hat{\rmR}_{i})}{\partial \hat{\rmR}_{i}}\biggl) ^\intercal &
\\ &=\left(\frac{\partial L(\hat{\rmR}_{i})}{\partial \hat{\rmR}_{i}}\rmB_i\rmQ \right)\Lambda\left( \rmQ^\intercal \rmB_i^\intercal\biggl(\frac{\partial L(\hat{\rmR}_{i})}{\partial \hat{\rmR}_{i}}\biggl) ^\intercal \right)
\\ & = \sum_k  \Lambda_{kk} \left( \rmQ^\intercal _{k} \rmB_i^\intercal\left(\frac{\partial L(\hat{\rmR}_{i})}{\partial \hat{\rmR}_{i}} \right)^\intercal \right)^{2}
\\ & \geq \lambda_{\text{min}}(\rmD\rmD^\intercal) \sum_k   \left( \rmQ^\intercal _{k}\rmB_i^\intercal\left(\frac{\partial L(\hat{\rmR}_{i})}{\partial \hat{\rmR}_{i}} \right)^\intercal\right)^{2}
 \\ & = \lambda_{\text{min}}(\rmD\rmD^\intercal)\left \| \rmQ^\intercal \rmB_i^\intercal\left(\frac{\partial L(\hat{\rmR}_{i})}{\partial \hat{\rmR}_{i}} \right)^\intercal\right\|^2
  \\ & = \lambda_{\text{min}}(\rmD\rmD^\intercal)\left \|\rmB_i^\intercal\left(\frac{\partial L(\hat{\rmR}_{i})}{\partial \hat{\rmR}_{i}} \right)^\intercal\right\|^2.
\end{align}
By expanding the squared Euclidean  norm,
\begin{align}
\epsilon^2 \ge\lambda_{\text{min}}(\rmD\rmD^\intercal)\frac{\partial L(\hat{\rmR}_{i})}{\partial \hat{\rmR}_{i}}(\rmB_{i}   \rmB_i^\intercal)\biggl(\frac{\partial L(\hat{\rmR}_{i})}{\partial \hat{\rmR}_{i}}\biggl) ^\intercal.
\end{align} 
Since  $\rmB_{i}   \rmB_i^\intercal$ is a real symmetric matrix, by repeating the same proof steps with eigendecomposition of $\rmB_{i}   \rmB_i^\intercal$, 
\begin{align}
    \epsilon^2 \geq 
    &   \lambda_{\text{min}}(\rmD\rmD^\intercal)  \lambda_{\text{min}}(\rmB_i\rmB_i^\intercal)\bigg\|\biggl(\frac{\partial L(\hat{\rmR}_{i})}{\partial \hat{\rmR}_{i}}\biggl)^\intercal\bigg\|^2.
\end{align}

\noindent
If $\lambda_{\text{min}}(\rmD\rmD^\intercal)  \lambda_{\text{min}}(\rmB_i\rmB_i)^\intercal\neq0$, by taking the square root of both sides, this implies that
\begin{equation}
    \bigg\|\biggl(\frac{\partial L(\hat{\rmR}_{i})}{\partial \hat{\rmR}_{i}}\biggl)^\intercal\bigg\| \leq\epsilon\psi(\rmD)\psi(\rmB_i).
\end{equation}

Since the training loss $\vw \rightarrow  L(\hat{\rmR}_{i})$ is non-convex while the loss criterion $\hat{\rmR}_{i} \rightarrow  L(\hat{\rmR}_{i})$ is convex, we have that
\begin{equation}
     L(\rmR^*) \geq  L(\hat{\rmR}_{i})+\frac{\partial L(\hat{\rmR}_{i})}{\partial \hat{\rmR}_{i}}(\rmR^*- \hat{\rmR}_{i}).
\end{equation}
\noindent
This implies that
\begin{align}
    L(\hat{\rmR}_{i}) &\leq L(\rmR^*) +\frac{\partial L(\hat{\rmR}_{i})}{\partial \hat{\rmR}_{i}}( \hat{\rmR}_{i}-\rmR^*)\\
    &\leq L(\rmR^*) + \bigg\|\biggl(\frac{\partial L(\hat{\rmR}_{i})}{\partial \hat{\rmR}_{i}}\biggl)^\intercal\bigg\|\|\hat{\rmR}_{i}-\rmR^*\| \\
    &\leq L(\rmR^*) + \epsilon E\psi(\rmD)\psi(\rmB_i).
\end{align}

\subsection{Proof of Theorem 2}
\label{app:proof_theorem_2}
\emph{Proof.} Since $\rmB_0\!=\!\frac{\partial\hat{\rmR}_{0}}{\partial\hat{\rmR}_{0}}\!=\!\rmI$, we have $\psi(\rmB_0)=1/\sqrt{\lambda_{\min}(\rmB_0\rmB_0\T)}\!=1/\sqrt{\lambda_{\min}(I)}\!=\!1$. This proves the statement for $\psi(\rmB_0)$. For the statement of $\psi(\rmB_1)$, we invoke a part of the inverse function theorem: i.e., if  the determinant of the Jacobian matrix of $g$ at its current input $\hat{\rmR}_0$ is nonzero, then $g$ is locally injective at  $\hat{\rmR}_{0}$. This implies that if $g$ is not locally injective at  $\hat{\rmR}_{0}$, then  the determinant of the Jacobian matrix of $g$ at  $\hat{\rmR}_0$ is zero. By noticing that  the Jacobian of $g$ is $\rmB_1$, this implies that $\lambda_{\min}(\rmB_1\rmB_1\T)=0$ and hence  $\psi(\rmB_1)=\infty$.

\subsection{On non-injectivity of orthonormalization} \label{app:1}
As a simple illustration, let $g(\rmA)$ represent a Gram--Schmidt orthonormalization of  a given matrix $\rmA=[v_{1},v_2,\dots,v_n]$. \ Then, we  can always perturb the second column of  $\rmA$ by $\rmA +\epsilon\Delta$ such that $g(\rmA)=g(\rmA+\epsilon\Delta)$ for any sufficiently small $\epsilon$ by setting $\Delta=[0,\delta,0,\dots,0]$ for any  $\delta$ satisfying  $\delta\T v_1 =0$. Thus, an orthogonalization map $g$ is inherently not locally injective and it is shown to cause the issue for  convergence to optimal matrices in Theorems 1--2. Thus, Theorems 1--2 show the advantages of the proposed method in terms of optimization. Future works include the theoretical analysis of generalization using information bottleneck \citep{icml2023kzxinfodl}. 

\subsection{Proof for downstream tasks}
\label{app:proof_downstream}

Similar to the above proof, we can arrive at 
\begin{equation}
    L(\hat{\rmY}_{i}) \leq L(\rmY^*) + \epsilon E\psi(\rmD)\psi(\rmW)\psi(\rmB_i), \,\,\,\text{where} \,\,\, \rmW = \frac{\partial \hat{\rmY}}{\partial\hat{\rmR}_i}.
\end{equation}

Since the downstream tasks usually consist of linear operations, it's easy to obtain that $\psi(\rmW)\!\neq\!0$.

\begin{figure}
    \centering
    \includegraphics[width=0.75\textwidth]{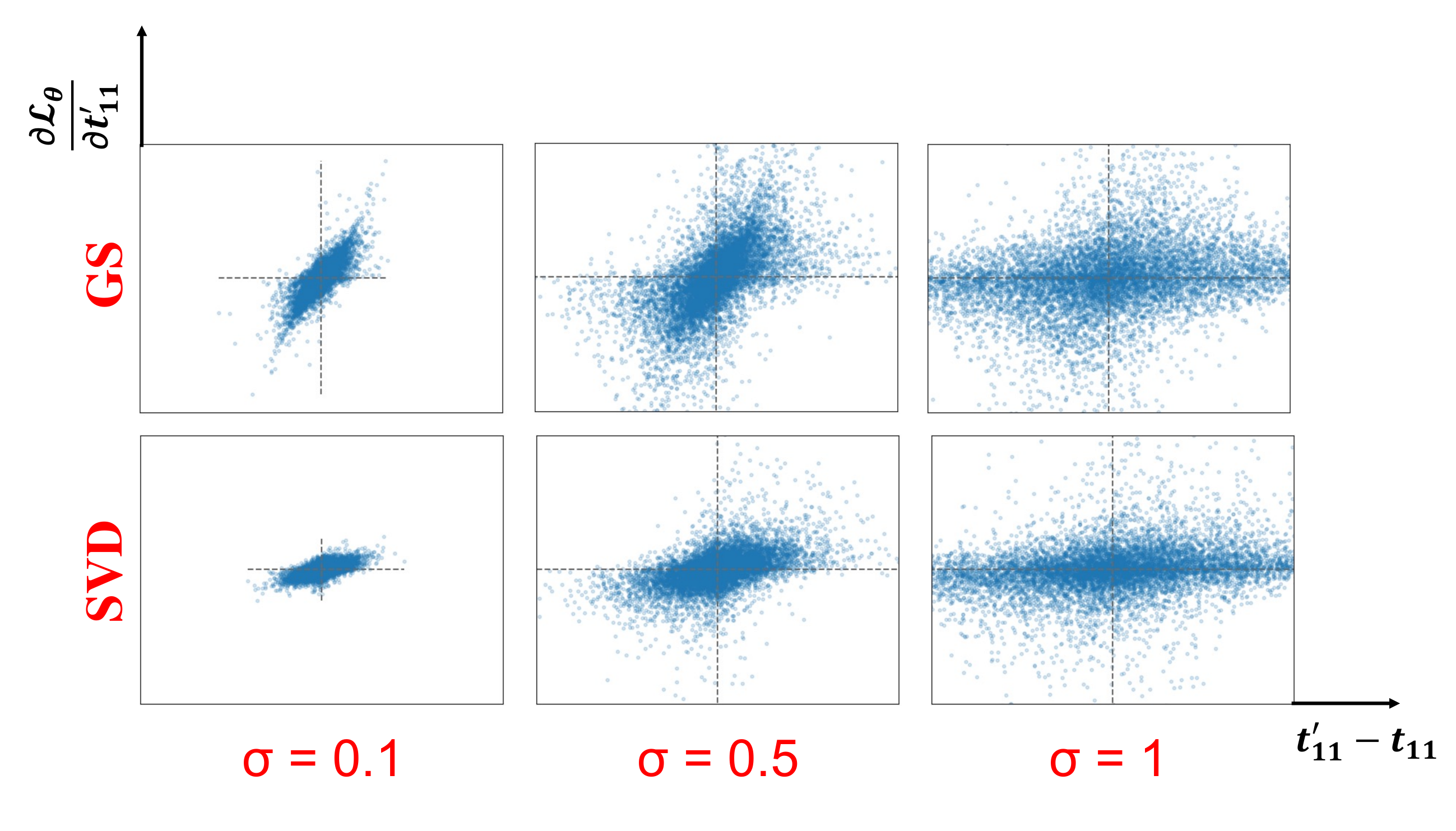}
    \caption{The ambiguity of incorporating Gram-Schmidt (GS) and SVD exists under each $\sigma$. With a larger $\sigma$, the gradient for the same $x$ becomes more diverse which indicates that at the beginning stage of training, the ambiguous gradients are severe and we claim this will influence the training efficiency.}
    \label{fig:full_gradient_sigma}
\end{figure}

\begin{figure}
    \centering
    \includegraphics[width=0.96\textwidth]{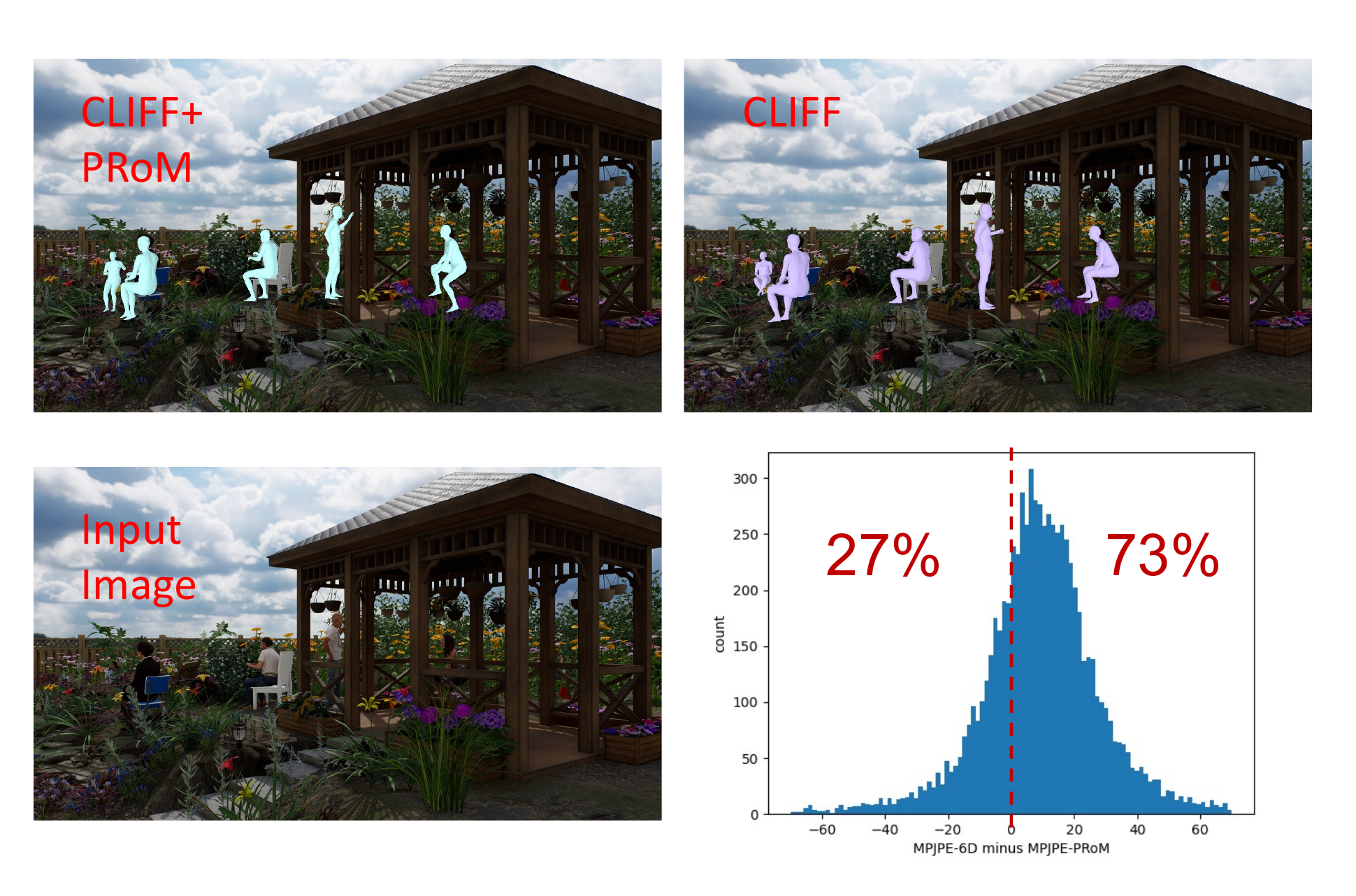}
    \caption{Qualitative results of PRoM vs 6D-based rotation representation on AGORA. We can see that PRoM achieves general improvement in samples. Statistically, with the same backbone, PRoM outperforms 6D-based representation on over 70\% of the total samples.}
    \label{fig:qualitative_results}
\end{figure}

\begin{table}[t]
\small
\def\arraystretch{1.0}
    \centering
    \resizebox{0.65\columnwidth}{!}{
    \begin{tabular}{l|ccc}
    \toprule
         &  MPJPE $\downarrow$ & PA-MPJPE $\downarrow$ & MPVPE $\downarrow$\\
     \midrule
     HMR~\cite{kanazawa2018end} & 74.4  & 46.6 & 87.3 \\
     HMR + PRoM & \textbf{71.4} & \textbf{44.5} & \textbf{84.6} \\
     \midrule
     PARE~\cite{kocabas2021pare} & 82.9 & 52.3& 99.7\\
     PARE + PRoM & \textbf{80.5}  & \textbf{49.7} & 96.8\\
     \midrule
     CLIFF (Res50)~\cite{li2022cliff} & 72.0 & 45.7 & 85.3 \\
     CLIFF (Res50) + PRoM & \textbf{70.8}  & \textbf{44.5} & \textbf{84.1}\\
     
     \bottomrule
    \end{tabular}}
    \caption{Evaluation of different models on 3DPW. '+PRoM' means replacing the 6D representation with a Pseudo Rotation Matrix representation.}
    \label{tab:ablation_model}
    \vspace{-0.5em}
\end{table}

\subsection{update ambiguity with different $\sigma$s}
\label{app:full_gradient_ambiguity}

To fully understand the update ambiguity under different noises, we present the visualization under three different $\sigma$s: 0.1, 0.5, and 1 as shown in Fig.~\ref{fig:full_gradient_sigma}. We can see that with the increase in variation, the gradients become more diverse.

\subsection{Details of Experimental Settings}
\label{app:details_experiments}

\textbf{Point Cloud Pose Estimation.} For the task of point cloud pose estimation, different rotation representations are estimated to rotate the reference point cloud $\rmP_r$ to a target point cloud $\rmP_t$. The network is required to directly output estimated rotations. During training, the loss for the point cloud pose estimation task $\mathcal{L}_{\text{pc}}$ is
\begin{equation}
    \mathcal{L}_{\text{pc}} = \mathcal{L}_\theta + \mathcal{L}_{\text{ele}}(\hat{\rmP}_t, \rmP_t).
\end{equation}
Different rotation representations have different $\hat{\theta}$s and $\theta$s. For axis-angle and Euler angle, $\theta\in\mathbb{R}^3$; for quaternions, $\theta\in\mathbb{R}^4$; for 6D representations, $\mathcal{L}_{\theta}\!=\!\mathcal{L}_{\text{ele}}(f_{\text{GS}}(\theta), \rmR)$ where $\theta\in\mathbb{R}^6$; for the proposed PRoM, $\theta\in\mathbb{R}^{9}$.

In summary, point cloud pose estimation is a task in which both rotation loss and downstream loss exist.

\textbf{3D Body/Hand Pose and Shape Estimation.} As illustrated in Sec.~\ref{app:details_downstream}, there are two downstream outputs, $\hat{\rmY}_M$ and $\hat{\rmY}_J$, which are directly associated with rotations. Since the previous paper~\citep{kolotouros2019learning} has shown the advantage of applying 6D representation over axis-angle. We only compare the proposed PRoM with 6D-representation-based methods. Therefore, the total loss for the task of body/hand pose and shape estimation is
\begin{equation}
    \mathcal{L}_{\text{ps}} = \mathcal{L}_\theta + \mathcal{L}_{\text{ele}}(\hat{\rmY}_M,\rmY_M) ++ \mathcal{L}_{\text{ele}}(\hat{\rmY}_J,\rmY_J).
\end{equation}
The above loss is applicable in Table~\ref{tab:sota},~\ref{tab:comb_2} (a),~\ref{tab:comb_2} (b) and~\ref{tab:ablation_model}. In summary, these experiments indicate the superiority of our method over 6D representation when both rotation loss and downstream loss exist.

\textbf{Pose and Shape Estimation with 2D Ground Truth.} In this setting, we have ground truth 2D keypoints as a weak supervision to refine the 3D predictions. Formally, by reprojecting the predicted 3D keypoints $\hat{\rmY}_J$ with the estimated camera, we apply an alignment loss between the reprojected 2D prediction and ground truth 2D locations:
\begin{equation}
    \mathcal{L}_{\text{ops}} = \mathcal{L}_{\text{ele}}(\hat{\rmY}_{J_{2D}}, \rmY_{J_{2D}})=\mathcal{L}_{\text{ele}}(\text{reproj}(\hat{\rmY}_J), \rmY_{J_{2D}}).
\end{equation}
Therefore, in this experiment, we demonstrate that when only downstream loss exists, our method still outperforms 6D representations.

\subsection{Implementation details of Body pose and shape estimation task}
\label{app:details_body_pse}
To compare with the state-of-the-art methods on 3D human body and hand mesh recovery, we take the recently introduced CLIFF~\cite{li2022cliff} as the baseline and replace the 6D representation with our proposed pseudo rotation matrices (PRoM). 
`CLIFF + PRoM' predicts in total $J\times9$ pose parameters, where $J$ is the joint number and each joint has a 9D prediction of PRoM.
During training, PRoM is used to calculate the pose loss and is fed to the parametric model (SMPL \cite{loper2015smpl} for body and MANO \cite{romero2022embodied} for hand) to calculate the 3D joint loss and 2D reprojection loss, instead of the rotation matrices from the Gram-Schmidt process.
The training setting is the same as CLIFF, except for the initial learning rate of 3e-4, which will cause gradient explosion in the original 6D version. For the experiments on AGORA~\cite{agora}, we use ViTPose~\cite{xu2022vitpose} as the backbone.
More specifically, for body pose, we use the Adam optimizer to train the model for 244K steps with a batch size of 256.
The learning rate is reduced by a factor of 10 at the $122\text{K}^{\text{th}}$ step.
The input images are cropped using the ground-truth bounding boxes, and resized to $256 \times 192$, preserving the aspect ratio.
For hand pose, we resize the cropped images to $224 \times 224$, train the model for 101K steps with a batch size of 128, and reduce the learning rate by a factor of 10 at the $70\text{K}^{\text{th}}$ and $90\text{K}^{\text{th}}$ steps.
During inference, we adopt Unbiased Gram-Schimidt process on PRoM to obtain the final rotation matrices.

\subsection{Visualizations of PRoM vs 6D-based methods}
\label{app:vis}
We give visualizations using the same backbone but with PRoM and 6D representation on the challenging dataset AGORA in Fig.~\ref{fig:qualitative_results}. From the bottom-right figure, we can conclude that PRoM has a general improvement over all samples than 6D-based methods.

\subsection{Details about Ablation of Different Models}
\label{app:details_different_models}
We demonstrate the ablation results in Table~\ref{tab:ablation_model}, which show significant improvement over all models. For the implementation details, we adopt the SOTA experiment setting to train the models with a mixture of 3D datasets (Human3.6M \cite{ionescu2013human3}, MPI-INF-3DHP \cite{mehta2017monocular} and 3DPW \cite{von2018recovering}) and 2D datasets (COCO \cite{lin2014microsoft} and MPII \cite{andriluka20142d}). 
The 3D pseudo-ground-truth for 2D datasets is provided by the CLIFF annotator \cite{li2022cliff}, except that we use the EFT pseudo-GT \cite{joo2021exemplar} for PARE-based models.
The image encoders are based on ResNet-50 \cite{ResNet}.

\end{document}

%% file: math_commands.tex
\usepackage{amsmath,amsfonts,bm}

\def\eqref#1{equation~\ref{#1}}

\def\1{\bm{1}}

\def\rmA{{\mathbf{A}}}
\def\rmB{{\mathbf{B}}}

\def\rmD{{\mathbf{D}}}

\def\rmI{{\mathbf{I}}}

\def\rmK{{\mathbf{K}}}

\def\rmM{{\mathbf{M}}}

\def\rmP{{\mathbf{P}}}
\def\rmQ{{\mathbf{Q}}}
\def\rmR{{\mathbf{R}}}

\def\rmU{{\mathbf{U}}}
\def\rmV{{\mathbf{V}}}
\def\rmW{{\mathbf{W}}}
\def\rmX{{\mathbf{X}}}
\def\rmY{{\mathbf{Y}}}

\def\vp{{\bm{p}}}
\def\vq{{\bm{q}}}
\def\vr{{\bm{r}}}

\def\vt{{\bm{t}}}

\def\vw{{\bm{w}}}

\DeclareMathAlphabet{\mathsfit}{\encodingdefault}{\sfdefault}{m}{sl}
\SetMathAlphabet{\mathsfit}{bold}{\encodingdefault}{\sfdefault}{bx}{n}













%% file: iclr2024_conference.bbl
\begin{thebibliography}{39}
\providecommand{\natexlab}[1]{#1}
\providecommand{\url}[1]{\texttt{#1}}
\expandafter\ifx\csname urlstyle\endcsname\relax
  \providecommand{\doi}[1]{doi: #1}\else
  \providecommand{\doi}{doi: \begingroup \urlstyle{rm}\Url}\fi

\bibitem[Andriluka et~al.(2014)Andriluka, Pishchulin, Gehler, and Schiele]{andriluka20142d}
Mykhaylo Andriluka, Leonid Pishchulin, Peter Gehler, and Bernt Schiele.
\newblock 2d human pose estimation: New benchmark and state of the art analysis.
\newblock In \emph{CVPR}, pp.\  3686--3693, 2014.

\bibitem[Bertsekas(1997)]{bertsekas1997nonlinear}
Dimitri~P Bertsekas.
\newblock Nonlinear programming.
\newblock \emph{Journal of the Operational Research Society}, 48\penalty0 (3):\penalty0 334--334, 1997.

\bibitem[Chang et~al.(2015)Chang, Funkhouser, Guibas, Hanrahan, Huang, Li, Savarese, Savva, Song, Su, et~al.]{chang2015shapenet}
Angel~X Chang, Thomas Funkhouser, Leonidas Guibas, Pat Hanrahan, Qixing Huang, Zimo Li, Silvio Savarese, Manolis Savva, Shuran Song, Hao Su, et~al.
\newblock Shapenet: An information-rich 3d model repository.
\newblock \emph{arXiv preprint arXiv:1512.03012}, 2015.

\bibitem[Choi et~al.(2021)Choi, Moon, Chang, and Lee]{choi2021beyond}
Hongsuk Choi, Gyeongsik Moon, Ju~Yong Chang, and Kyoung~Mu Lee.
\newblock Beyond static features for temporally consistent 3d human pose and shape from a video.
\newblock In \emph{CVPR}, pp.\  1964--1973, 2021.

\bibitem[Fehrman et~al.(2020)Fehrman, Gess, and Jentzen]{fehrman2020convergence}
Benjamin Fehrman, Benjamin Gess, and Arnulf Jentzen.
\newblock Convergence rates for the stochastic gradient descent method for non-convex objective functions.
\newblock \emph{The Journal of Machine Learning Research}, 21\penalty0 (1):\penalty0 5354--5401, 2020.

\bibitem[Grassia(1998)]{grassia1998practical}
F~Sebastian Grassia.
\newblock Practical parameterization of rotations using the exponential map.
\newblock \emph{Journal of graphics tools}, 3\penalty0 (3):\penalty0 29--48, 1998.

\bibitem[Hasson et~al.(2019)Hasson, Varol, Tzionas, Kalevatykh, Black, Laptev, and Schmid]{hasson2019learning}
Yana Hasson, Gul Varol, Dimitrios Tzionas, Igor Kalevatykh, Michael~J Black, Ivan Laptev, and Cordelia Schmid.
\newblock Learning joint reconstruction of hands and manipulated objects.
\newblock In \emph{CVPR}, pp.\  11807--11816, 2019.

\bibitem[He et~al.(2016)He, Zhang, Ren, and Sun]{ResNet}
Kaiming He, Xiangyu Zhang, Shaoqing Ren, and Jian Sun.
\newblock Deep residual learning for image recognition.
\newblock In \emph{CVPR}, pp.\  770--778, 2016.

\bibitem[Ionescu et~al.(2013)Ionescu, Papava, Olaru, and Sminchisescu]{ionescu2013human3}
Catalin Ionescu, Dragos Papava, Vlad Olaru, and Cristian Sminchisescu.
\newblock Human3. 6m: Large scale datasets and predictive methods for 3d human sensing in natural environments.
\newblock \emph{IEEE TPAMI}, 36\penalty0 (7):\penalty0 1325--1339, 2013.

\bibitem[Ionescu et~al.(2015)Ionescu, Vantzos, and Sminchisescu]{ionescu2015matrix}
Catalin Ionescu, Orestis Vantzos, and Cristian Sminchisescu.
\newblock Matrix backpropagation for deep networks with structured layers.
\newblock In \emph{ICCV}, pp.\  2965--2973, 2015.

\bibitem[Joo et~al.(2021)Joo, Neverova, and Vedaldi]{joo2021exemplar}
Hanbyul Joo, Natalia Neverova, and Andrea Vedaldi.
\newblock Exemplar fine-tuning for 3d human model fitting towards in-the-wild 3d human pose estimation.
\newblock In \emph{3DV}, pp.\  42--52. IEEE, 2021.

\bibitem[Kanazawa et~al.(2018)Kanazawa, Black, Jacobs, and Malik]{kanazawa2018end}
Angjoo Kanazawa, Michael~J Black, David~W Jacobs, and Jitendra Malik.
\newblock End-to-end recovery of human shape and pose.
\newblock In \emph{CVPR}, pp.\  7122--7131, 2018.

\bibitem[Kawaguchi \& Bengio(2019)Kawaguchi and Bengio]{kawaguchi2019depth}
Kenji Kawaguchi and Yoshua Bengio.
\newblock Depth with nonlinearity creates no bad local minima in resnets.
\newblock \emph{Neural Networks}, 118:\penalty0 167--174, 2019.

\bibitem[Kawaguchi et~al.(2022)Kawaguchi, Zhang, and Deng]{kawaguchi2022understanding}
Kenji Kawaguchi, Linjun Zhang, and Zhun Deng.
\newblock Understanding dynamics of nonlinear representation learning and its application.
\newblock \emph{Neural computation}, 34\penalty0 (4):\penalty0 991--1018, 2022.

\bibitem[Kawaguchi et~al.(2023)Kawaguchi, Deng, Ji, and Huang]{icml2023kzxinfodl}
Kenji Kawaguchi, Zhun Deng, Xu~Ji, and Jiaoyang Huang.
\newblock How does information bottleneck help deep learning?
\newblock In \emph{International Conference on Machine Learning (ICML)}, 2023.

\bibitem[Knutsson et~al.(2011)Knutsson, Westin, and Andersson]{knutsson2011representing}
Hans Knutsson, Carl-Fredrik Westin, and Mats Andersson.
\newblock Representing local structure using tensors ii.
\newblock In \emph{Scandinavian conference on image analysis}, pp.\  545--556. Springer, 2011.

\bibitem[Kocabas et~al.(2020)Kocabas, Athanasiou, and Black]{kocabas2020vibe}
Muhammed Kocabas, Nikos Athanasiou, and Michael~J Black.
\newblock Vibe: Video inference for human body pose and shape estimation.
\newblock In \emph{CVPR}, pp.\  5253--5263, 2020.

\bibitem[Kocabas et~al.(2021)Kocabas, Huang, Hilliges, and Black]{kocabas2021pare}
Muhammed Kocabas, Chun-Hao~P Huang, Otmar Hilliges, and Michael~J Black.
\newblock Pare: Part attention regressor for 3d human body estimation.
\newblock In \emph{ICCV}, 2021.

\bibitem[Kolotouros et~al.(2019)Kolotouros, Pavlakos, Black, and Daniilidis]{kolotouros2019learning}
Nikos Kolotouros, Georgios Pavlakos, Michael~J Black, and Kostas Daniilidis.
\newblock Learning to reconstruct 3d human pose and shape via model-fitting in the loop.
\newblock In \emph{ICCV}, pp.\  2252--2261, 2019.

\bibitem[Lei et~al.(2019)Lei, Hu, Li, and Tang]{lei2019stochastic}
Yunwen Lei, Ting Hu, Guiying Li, and Ke~Tang.
\newblock Stochastic gradient descent for nonconvex learning without bounded gradient assumptions.
\newblock \emph{IEEE transactions on neural networks and learning systems}, 31\penalty0 (10):\penalty0 4394--4400, 2019.

\bibitem[Levinson et~al.(2020)Levinson, Esteves, Chen, Snavely, Kanazawa, Rostamizadeh, and Makadia]{levinson2020analysis}
Jake Levinson, Carlos Esteves, Kefan Chen, Noah Snavely, Angjoo Kanazawa, Afshin Rostamizadeh, and Ameesh Makadia.
\newblock An analysis of svd for deep rotation estimation.
\newblock \emph{NeurIPS}, 33:\penalty0 22554--22565, 2020.

\bibitem[Li et~al.(2021)Li, Xu, Chen, Bian, Yang, and Lu]{li2021hybrik}
Jiefeng Li, Chao Xu, Zhicun Chen, Siyuan Bian, Lixin Yang, and Cewu Lu.
\newblock Hybrik: A hybrid analytical-neural inverse kinematics solution for 3d human pose and shape estimation.
\newblock In \emph{CVPR}, pp.\  3383--3393, 2021.

\bibitem[Li et~al.(2022)Li, Liu, Zhang, Xu, and Yan]{li2022cliff}
Zhihao Li, Jianzhuang Liu, Zhensong Zhang, Songcen Xu, and Youliang Yan.
\newblock Cliff: Carrying location information in full frames into human pose and shape estimation.
\newblock In \emph{ECCV}, 2022.

\bibitem[Lin et~al.(2021)Lin, Wang, and Liu]{lin2021end}
Kevin Lin, Lijuan Wang, and Zicheng Liu.
\newblock End-to-end human pose and mesh reconstruction with transformers.
\newblock In \emph{CVPR}, pp.\  1954--1963, 2021.

\bibitem[Lin et~al.(2014)Lin, Maire, Belongie, Hays, Perona, Ramanan, Doll{\'a}r, and Zitnick]{lin2014microsoft}
Tsung-Yi Lin, Michael Maire, Serge Belongie, James Hays, Pietro Perona, Deva Ramanan, Piotr Doll{\'a}r, and C~Lawrence Zitnick.
\newblock Microsoft {COCO}: Common objects in context.
\newblock In \emph{ECCV}, pp.\  740--755. Springer, 2014.

\bibitem[Loper et~al.(2015)Loper, Mahmood, Romero, Pons-Moll, and Black]{loper2015smpl}
Matthew Loper, Naureen Mahmood, Javier Romero, Gerard Pons-Moll, and Michael~J Black.
\newblock Smpl: A skinned multi-person linear model.
\newblock \emph{ACM TOG}, 34\penalty0 (6):\penalty0 1--16, 2015.

\bibitem[Mehta et~al.(2017)Mehta, Rhodin, Casas, Fua, Sotnychenko, Xu, and Theobalt]{mehta2017monocular}
Dushyant Mehta, Helge Rhodin, Dan Casas, Pascal Fua, Oleksandr Sotnychenko, Weipeng Xu, and Christian Theobalt.
\newblock Monocular 3d human pose estimation in the wild using improved cnn supervision.
\newblock In \emph{3DV}, pp.\  506--516. IEEE, 2017.

\bibitem[Mertikopoulos et~al.(2020)Mertikopoulos, Hallak, Kavis, and Cevher]{mertikopoulos2020almost}
Panayotis Mertikopoulos, Nadav Hallak, Ali Kavis, and Volkan Cevher.
\newblock On the almost sure convergence of stochastic gradient descent in non-convex problems.
\newblock \emph{Advances in Neural Information Processing Systems}, 33:\penalty0 1117--1128, 2020.

\bibitem[Moon \& Lee(2020)Moon and Lee]{moon2020i2l}
Gyeongsik Moon and Kyoung~Mu Lee.
\newblock I2l-meshnet: Image-to-lixel prediction network for accurate 3d human pose and mesh estimation from a single rgb image.
\newblock In \emph{ECCV}, pp.\  752--768. Springer, 2020.

\bibitem[Moon et~al.(2020)Moon, Yu, Wen, Shiratori, and Lee]{moon2020interhand2}
Gyeongsik Moon, Shoou-I Yu, He~Wen, Takaaki Shiratori, and Kyoung~Mu Lee.
\newblock Interhand2. 6m: A dataset and baseline for 3d interacting hand pose estimation from a single rgb image.
\newblock In \emph{ECCV}, pp.\  548--564. Springer, 2020.

\bibitem[Patel et~al.(2021)Patel, Huang, Tesch, Hoffmann, Tripathi, and Black]{agora}
Priyanka Patel, Chun-Hao~P. Huang, Joachim Tesch, David~T. Hoffmann, Shashank Tripathi, and Michael~J. Black.
\newblock {AGORA}: Avatars in geography optimized for regression analysis.
\newblock In \emph{CVPR}, 2021.

\bibitem[Rieger \& Van~Vliet(2004)Rieger and Van~Vliet]{rieger2004systematic}
Bernd Rieger and Lucas~J Van~Vliet.
\newblock A systematic approach to nd orientation representation.
\newblock \emph{Image and Vision Computing}, 22\penalty0 (6):\penalty0 453--459, 2004.

\bibitem[Romero et~al.(2017)Romero, Tzionas, and Black]{romero2022embodied}
Javier Romero, Dimitrios Tzionas, and Michael~J Black.
\newblock Embodied hands: Modeling and capturing hands and bodies together.
\newblock \emph{SIGGRAPH Asia}, 2017.

\bibitem[Saxena et~al.(2009)Saxena, Driemeyer, and Ng]{saxena09}
A.~Saxena, J.~Driemeyer, and A.~Ng.
\newblock Learning 3-d object orientation from images.
\newblock In \emph{ICRA}, 2009.

\bibitem[Von~Marcard et~al.(2018)Von~Marcard, Henschel, Black, Rosenhahn, and Pons-Moll]{von2018recovering}
Timo Von~Marcard, Roberto Henschel, Michael~J Black, Bodo Rosenhahn, and Gerard Pons-Moll.
\newblock Recovering accurate 3d human pose in the wild using imus and a moving camera.
\newblock In \emph{ECCV}, pp.\  601--617, 2018.

\bibitem[Wang et~al.(2021)Wang, Dang, Hu, Fua, and Salzmann]{wang2021robust}
Wei Wang, Zheng Dang, Yinlin Hu, Pascal Fua, and Mathieu Salzmann.
\newblock Robust differentiable svd.
\newblock \emph{IEEE TPAMI}, 2021.

\bibitem[Xu et~al.(2022)Xu, Zhang, Zhang, and Tao]{xu2022vitpose}
Yufei Xu, Jing Zhang, Qiming Zhang, and Dacheng Tao.
\newblock Vitpose: Simple vision transformer baselines for human pose estimation.
\newblock \emph{arXiv preprint arXiv:2204.12484}, 2022.

\bibitem[Zhou et~al.(2019)Zhou, Barnes, Lu, Yang, and Li]{zhou2019continuity}
Yi~Zhou, Connelly Barnes, Jingwan Lu, Jimei Yang, and Hao Li.
\newblock On the continuity of rotation representations in neural networks.
\newblock In \emph{CVPR}, 2019.

\bibitem[Zimmermann et~al.(2019)Zimmermann, Ceylan, Yang, Russell, Argus, and Brox]{zimmermann2019freihand}
Christian Zimmermann, Duygu Ceylan, Jimei Yang, Bryan Russell, Max Argus, and Thomas Brox.
\newblock Freihand: A dataset for markerless capture of hand pose and shape from single rgb images.
\newblock In \emph{ICCV}, pp.\  813--822, 2019.

\end{thebibliography}
